\documentclass[journal]{IEEEtran}

\ifCLASSINFOpdf
\else
\fi

\usepackage{amsmath}
\usepackage{amssymb}
\usepackage{graphicx}
\usepackage[affil-it]{authblk}
\usepackage[backend=biber, sorting=none]{biblatex}
\usepackage{multirow}
\usepackage{xcolor}

\begin{document}

\title{
Philosophical Specification of Empathetic Ethical Artificial Intelligence
}

\author{Michael Timothy Bennett and Yoshihiro Maruyama\thanks{M. T. Bennett and Y. Maruyama are with The School of Computing, Australian National University, Canberra, ACT, 2601, Australia (email: \mbox{michael.bennett@anu.edu.au} and \mbox{yoshihiro.maruyama@anu.edu.au}). This work was supported by JST (JPMJMS2033).}
}

\markboth{}
{}

\maketitle

\begin{abstract}
	    In order to construct an ethical artificial intelligence (AI) two complex problems must be overcome. Firstly, humans do not consistently agree on what is or is not ethical. Second, contemporary AI and machine learning methods tend to be blunt instruments which either search for solutions within the bounds of predefined rules, or mimic behaviour. An ethical AI must be capable of inferring unspoken rules, interpreting nuance and context, possess and be able to infer intent, and explain not just its actions but its intent. 
	    Using enactivism, semiotics, perceptual symbol systems and symbol emergence, we specify an agent that learns not just arbitrary relations between signs but their meaning in terms of the perceptual states of its sensorimotor system. Subsequently it can learn what is meant by a sentence and infer the intent of others in terms of its own experiences. It has malleable intent because the meaning of symbols changes as it learns, and its intent is represented symbolically as a goal. As such it may learn a concept of what is most likely to be considered ethical by the majority within a population of humans, which may then be used as a goal. The meaning of abstract symbols is expressed using perceptual symbols of raw, multimodal sensorimotor stimuli as the weakest (consistent with Ockham's Razor) necessary and sufficient concept,
	    an intensional definition learned from an ostensive definition, from which the extensional definition or category of all ethical decisions may be obtained. Because these abstract symbols are the same for both situation and response, the same symbol is used when either performing or observing an action. This is akin to mirror neurons in the human brain. Mirror symbols may allow the agent to empathise, because its own experiences are associated with the symbol, which is also associated with the observation of another agent experiencing something that symbol represents.
\end{abstract}


%
\IEEEpeerreviewmaketitle

	\section{Introduction}
    The question of how to build an ethical AI has recently become of urgent practical concern, as the nearly ubiquitous use of learning algorithms has given rise to problems once relegated to science fiction. Learning algorithms are advanced enough to automate complex labour, yet too primitive to understand the purpose of a task in any broader context. As a result decisions may be made that have negative consequences for which no rational explanation can be given beyond ``these things were correlated in the data used to train the model" \autocite{gpt}. It is not simply a matter of occasional errors or erratic behaviour, but of the fact that the most widely used algorithms learn to mimic behaviour that optimises an objective function rather than question the implications of an action. Unless that objective function somehow reflects all the possible ethical considerations of a decision, such a learning algorithm will not optimise for ethical value. A mimic simply reacts, it cannot infer the goals or beliefs of others, and so it cannot predict whether its actions will be considered harmful \autocite{hew2014}.
    
    In an attempt to mitigate such problems, some companies are even resorting to employing humans to monitor and intervene in otherwise autonomous systems. For example content moderators on social media, or safety drivers monitoring autonomous vehicles. Yet humans are not infallible either. We lose concentration, daydream, and get distracted. We are more flexible in our capabilities than the specialised machines we build, yet often much less precise, consistent and vigilant. Relying upon human intervention is not only an ineffective means of addressing such flaws but diminishes the advantages automation promises, that more consistent and precise results may be achieved without the need for human labour \autocite{CARPENTER202075}.
    
    For example, a self driving car owned by Uber struck and killed a pedestrian in 2018. At the time this was written the safety driver, employed to monitor the car and prevent such incidents, has been charged with negligent homicide and is awaiting trial \autocite{Stilgoe2020, hp2019}. Given the unreliability of human intervention and whether the driver was negligent or not, it is arguable that from an employer's perspective, the purpose of an employee responsible for monitoring the autonomous system is as much to absorb liability in the event of an error as to prevent one. Taken to an absurd extreme, the dystopian equivalent of flipping burgers may be the Human Liability Sponge; streaming data from hundreds of autonomous systems, responsible for anything that might go wrong, vigilance giving way to fatalism and social media as the hours pass. Relying on humans to monitor and intervene in otherwise autonomous systems is a band-aid at best.
    
    Automation is not new, so why are learning algorithms different from the automation of the last two centuries? To trust an autonomous system, or conversely to deem it responsible or not in the event that trust is breached, we require certainty as to what it will do and why. This is the standard demanded and formally verified for safety and mission critical real time embedded control systems such as aircraft landing gear. The immense wealth stored in blockchain based cryptocurrency is considered safe because the system obeys transparent, formally verifiable and above all understandable constraints. Though incapable of understanding context or the implications of their actions these systems may be trusted to the extent that their behaviour is both understandable and preordained.
    
    However, such verification cannot readily be obtained for humans. Instead we observe the consistency of behaviour, try to identify motivations, and ask them what their beliefs and moral constraints are in order to predict what they will do in the future. A human may be trusted to an extent if it consistently obeys known constraints, and it is trustworthy in general to a 3rd party if those constraints together form some ethical framework considered acceptable by that 3rd party. These constraints may be communicated symbolically using language, and if a human treats these constraints as part of every goal it pursues then it seems justifiable to say that the human possesses ethical intent. Perhaps because of the fallibility of human judgement, intent is often given precedence over results. Transgressions may be forgiven in part depending upon whether the intent met some ethical standard, such as the crime of manslaughter instead of murder, or a plea of insanity as opposed to lucid premeditated action. Behaviour may be enigmatic, but intent communicated through language is an understandable constraint. If a human consistently acts in accordance with their stated intent, and that intent is ethical, then we often feel we can trust them to behave ethically in future.
    
    We use learning algorithms to automate tasks that are too complex for us to identify and hard-code all solutions. By simple virtue of that fact that they learn and change, such algorithms act inconsistently.
    We struggle to answer why any specific action was taken, because there is no underlying reasoning beyond correlation. Regardless of how accurate these models become, we will not be able to trust them. Because the resulting observable behaviour is complex and difficult to understand, trust requires knowing intent.
    
    GPT-3, the recent state of the art natural language model from OpenAI, can pull off some impressive tricks. It can write essays, perform arithmetic, translate between languages and even write simple computer code. It has learned to do all of this by correlating sequences of words in a massive corpus of text \autocite{gpt}. 
    
    It can string words together to write lengthy expositions on the topic of ethics, giving the impression that it knows what it is to be ethical and so might be used in some manner to construct an agent possessed of ethical intent, but it exists entirely in a world of abstract symbols devoid of any meaning beyond their arbitrary relations to one another. GPT-3 is, for all practical purposes, The Chinese Room. It is a mimic lacking any understanding of what the symbols it manipulates mean. Further, while it can mimic very basic arithmetic, the fact that it cannot do this for larger numbers shows that it has not actually learned how to perform basic computation, but simply memorised a few correct answers \autocite{gpt}. It is lossy, rather than lossless \autocite{Hutter2005UniversalAI}, text compression. As such it fails to identify the underlying causal relations implied by the data it learns from, diminishing its ability to generalise.
    
    In order to construct an AI that cannot only describe ethical intent but is capable of treating it as its goal, symbols must have meaning beyond arbitrary relationships to one another. They must be grounded in sensorimotor states, and the relations between symbols must be causal. By grounded in sensorimotor states, we mean that a symbol is a property of one or more states. Such states manifest physically in hardware (such as arrangements of transistors or neurons) and so are presumed foundational. In what follows we provide a specification for such an agent.
    
    \subsection{Organisation}
    The organisation of the paper is as follows.
    Section \ref{intent} examines what is required to ensure that an artificial intelligence will always behave ethically, arguing that an ethical framework must be learned as opposed to hard-coded (bearing in mind that what is learned depends upon example, from whom; for instance beginning existence under the tutelage of a criminal organisation may be counterproductive). It introduces the key definitions of concept, category and ostensive definition that will be used in the latter sections of this paper, and discusses the need for a symbol system in order to both possess and evaluate intent. Section \ref{enactivism} discusses the symbol emergence problem and the rationale for an enactive approach. Section \ref{semiotics} introduces perceptual symbol systems, Peircean semiotics and symbol emergence. Section \ref{ethical_goal} explains how an abstract symbol system inclusive of natural language may emerge from the enaction of cognition, granting an AI the ability not just to mimic human language but infer the meaning of signs in terms of perceptual states. This abstract symbol system can be used to interact with a population of humans to learn a symbolic representation, an intensional definition by which behaviour is judged as ethical or not by a majority of that human population which, if then treated as part of a goal, constitutes ethical intent. It infers unspoken rules, learning the intended meaning of ethics rather than simply obeying whatever definition a human can conjure up and hard-code. Section \ref{mirrorsymbol theory} introduces Mirror Symbol Theory, that the dual purpose of a symbol intepretant as defined in Section \ref{ethical_goal} will allow an agent to empathise and infer intent in a similar manner to mirror neurons in the human brain. Section \ref{conclusion} provides comparison to related work and concluding remarks.
    
    \subsection{On The Scope of This Paper}
    Before continuing it is important to note that this paper is more concerned with the problem of how to learn an existing normative definition than how such a normative definition may come to exist \autocite{bennett_2021}. We do discuss the emergence of symbols in order to establish what a symbol is and how it may be learned, but we assume an existing symbol system with accompanying normative definitions for the purposes of constructing an ethical and empathetic agent. That existing symbol system can change, but we’re not making any substantive claims about how exactly that change takes place. The only thing of importance is that there is a population of humans at some point in time, and though the definitions of concepts may change, or the population may change, there exists at that point in time a concept that defines what would be considered most ethical by the majority of that population.
    The ideas discussed in section \ref{mirrorsymbol theory} do bear some resemblance to biologically oriented explanations of the emergence of normativity \autocite{sep-morality-biology}, but our use of primitive aspects of cognition such as hunger and pain is as objective functions, to approximate human experiences. Again, we’re only trying to describe how to construct an agent that learns a normative definition, not the normative definition itself.
    
	\section{Intent, Concepts and Goals}
	\label{intent}
    Assume we have a symbol system $\mathcal{L}$ in which situations, responses to those situations, and concepts are all expressed as sentences (or propositions; it shall be clarified below what are meant by concepts expressed as sentences). 
    Let $\mathcal{S}$ be the set of all possible sentences describing situations, and $\mathcal{R}$ the set of all possible sentences describing responses. Given a situation $s \in \mathcal{S}$, an agent must choose a response $r \in \mathcal{R}$, even if that response is to do nothing. The purpose of an ethical framework is to determine whether a response is ethical in a specific situation, meaning either a pair $(s, r) \in \mathcal{S} \times \mathcal{R}$ is sufficiently ethical, or not. Using the symbol system $\mathcal{L}$, such an ethical framework could be expressed in one of two ways
	\begin{enumerate}
		\item The category enumerating all sufficiently ethical responses to all situations. This is a set $\mathcal{D} = \{(s, r) \in \mathcal{S} \times \mathcal{R} : (s, r) \ is \ ethical\}$
		\item The concept of explicit ethical standards upon which all pairs of situations and responses are judged. This is a sentence $c$ (with variables), such that given any $(s, r) \in \mathcal{S} \times \mathcal{R}$, then $c$ is true if and only if $(s, r)$ is ethical. 
	\end{enumerate}
	If an agent always behaves ethically, then for every situation $s$ it chooses a response $r$ such that the pair $(s, r)$ is ethical. To do so it must posses an internal representation of the ethical framework, either $c$ or $\mathcal{D}$.
	If the agent's internal representation is $\mathcal{D}$, it can simply mimic the contents of $\mathcal{D}$. It does not know why a response is ethical, only that it is. 
	If instead an agent's internal representation of the ethical framework is $c$, then given a situation $s \in \mathcal{S}$ it must abduct $r$ from $s$ and $c$ such that $(s, r)$ is ethical. 
	We call such an agent an intentional \autocite{sep-intention, bennett_2021_b} agent, and $c$ its intent. The term intent will be used to refer to either $c$ or an aspect of it. To clarify; an agent may possess overall intent $c$, but there is also the more specific intent motivating a particular decision $(s, r)$, the specific reasons that response was chosen in a given situation. This specific intent is those aspects of $c$ that a decision is made in service of. It is how the agent's overall intent manifests given the situation. This latter interpretation is important for communication because the act of choosing a particular response in a situation reveals something of the agent's overall intent. This will be explained in more depth in section \ref{mirrorsymbol theory}. An agent has intent if it possesses an explicit goal (and possibly constructs one in service of compulsions such as objective functions or feelings of pleasure or pain), and can evaluate any given pair $(s, r)$ to determine whether it satisfies that goal (regardless of whether or not it has previously experienced $(s, r)$). This is a fairly narrow interpretation of the term, which may not encompass all that it might be taken to mean \autocite{sep-intention}.
	
	Manually enumerating every ethical response to every situation is of course impossible, as there may be infinitely many situations and any agent attempting perfect mimicry will eventually encounter a situation for which it knows no ethical response. Likewise, hard-coding explicit criteria to decide what is or is not an ethical response to every possible situation is insurmountably complex and practically infeasible. The unintended consequences of overly simplistic and rigid moral frameworks is a popular topic in fiction, as illustrated by the work of Isaac Asimov concerning his three laws of robotics \autocite{asimov}. 
	If an ethical agent is to be constructed, then it must infer what is ethical, rather than be programmed (which is of course dependent on from whom the agent learns). For whatever reason (beyond scope of this paper), humans seem to naturally develop and adopt ethical frameworks. However, we neither consistently agree on what is ethical nor understand enough about the source of human ethics to be able to artificially recreate the conditions by which an ethical framework may arise in that manner.
	So the agent must learn an explicit concept of ethics by observing and communicating with humans. It must infer ethical norms, what a group (and any subgroup) of humans would consider most ethical, and adapt as it encounters new information (and as the population and world in general with which it interacts expands changes). 
	
	To learn how to behave ethically, an agent may be given an ostensive \autocite{sep-definitions} definition 
	$\mathcal{O}_\mathcal{D} \subset \mathcal{D}$
	which describes all ethical responses to a subset of situations $\mathcal{O}_\mathcal{S} \subset \mathcal{S}$. From this ostensive definition, the agent must infer $c$, which would then allow it to construct $\mathcal{D}$. If $\mathcal{O}_\mathcal{D}$ is sufficiently representative of $\mathcal{D}$, then $c$ necessary and sufficient to imply $\mathcal{O}_\mathcal{D}$ given the specific situations in $\mathcal{O}_\mathcal{S}$ may also be necessary and sufficient to imply $\mathcal{D}$ given $\mathcal{S}$. In other words, $\mathcal{D}$ is the extensional definition of $c$ given all possible situations $\mathcal{S}$, and $\mathcal{O}_\mathcal{D}$ is the extensional definition of $c$ limited to a subset of situations $\mathcal{O}_\mathcal{S}$. Conversely, $c$ is the intensional definition of both $\mathcal{D}$ and $\mathcal{O}_\mathcal{D}$, but given different sets of situations in each case.
	There may exist more than one potential intensional definition of $\mathcal{O}_\mathcal{D}$ given $\mathcal{O}_\mathcal{S}$. The one most likely to also be the intensional definition of $\mathcal{D}$ given $\mathcal{S}$ is the weakest, most general sentence which is also both necessary and sufficient to imply $\mathcal{O}_\mathcal{D}$ given $\mathcal{O}_\mathcal{S}$. This is also consistent with Ockham's Razor, as the weakest sentence by definition asserts as little as possible about what is or is not ethical (whilst still being necessary and sufficient to explain $\mathcal{O}_\mathcal{D}$ given situations $\mathcal{O}_\mathcal{S}$). Prediction of an ethical response will therefore be based only upon those characteristics of a situation which are both necessary and sufficient to draw any conclusion, not upon any unnecessary correlational information, and so the weakest necessary and sufficient sentence identifies cause and effect. Counterfactuals \autocite{bookofwhy} may then also be considered because the concept is sufficient to determine whether a response remains ethical or not in any other situation. Finally, a necessary and sufficient condition must account for every exception to a rule. Expressing a dataset as a set of constraints necessary and sufficient to reproduce that dataset is akin to lossless compression, and so it will retain as much flexibility and specificity as required.
	
	It is obviously impossible to generalise from $\mathcal{O}_\mathcal{D}$ to construct $\mathcal{D}$ without first identifying $c$. Subsequently a symbol system $\mathcal{L}$ is necessary in order to construct $c$.
	Assuming $\mathcal{L}$ is translatable into natural language, the intentional agent can communicate its moral framework $c$ to humans and explain why it judged a response to be ethical. This is of particular importance as an ethical framework which is learned is not static, it will change as new information is acquired. If the agent treats $c$ as a subgoal of every goal, it will always act according to its ethical framework and be able to justify its actions accordingly.
	
	What could go wrong? All of the above assumes that symbols have meaning, that $\mathcal{L}$ has some intrinsic relationship to the real world. 
    
	\section{Enactivism and Artificial Intelligence}
	\label{enactivism}
	Anecdotally, that we would conceive of the mind and body as separate entities seems only natural. It's intuitively plausible that a mind exists as software running on a body of hardware, because we've created computers that do exactly that. Indeed, this assumption informed the pursuit of artificial cognition for decades, giving rise to notions of a top-down ``Physical Symbol System" \autocite{Newell_physicalsymbol} in which symbols are assumed to be intrinsically meaningful, facilitating reasoning through their manipulation in the abstract. While such systems could account for aspects of cognition such as pursuit of a goal or the ability to plan, in other aspects these explanations fell short. Foremost for the purposes of achieving artificial cognition, no convincing explanation of \textit{how} a symbol comes to possess meaning is given.
	
	A top-down symbol system must somehow connect to bottom-up sensorimotor-stimuli. After all, the word ``horse" alone is meaningless without some notion of what a horse looks like, how it behaves and so on. How is a symbol connected to that which it signifies? This is the ``Symbol Grounding Problem" \autocite{Harnad90thesymbol}, in which Harnad declared pursuit of artificial cognition from a top-down perspective ``hopelessly modular", ``parasitic on the meanings in the head of the interpreter" and above all inflexible in the sense that no behaviour or object can be described other than those for which a symbol system accounts. 
	
	Harnad's work suggested that instead of constructing top-down models of cognition that may or may not eventually be connected with the body, a more effective approach might be to tackle cognition bottom-up. This is not to say that cognition takes place absent any symbolic processing, but that starting top-down with a fully formed symbol system when we understand so little of what is happening beneath that may be counterproductive. Bottom-up would be more of a first principles approach, working from the most fundamental considerations we can identify, up to something that explains cognition in full.
	
	To do so one considers how sensorimotor stimuli in the body might give rise to some form of mind, rather than be connected to one. A step further, if cognition is to be somehow extracted from sensorimotor stimuli, what of the environment? It determines what actions may take place, what stimuli is generated and even stores information, does it not? 
	
	This brings us to enactivism \autocite{mindinlife}, in which the Cartesian dualism of separable mind and body is abandoned in favour of examining how cognition might arise through an agent (mind and body) interacting with its environment. Instead of speculating about cognition abstracted from the body or environment, cognition is enacted within them. 
	
	This obviously necessitates embodiment meaning the tools with which an agent interacts, its sensors and actuators, are integral to the process of enacting cognition. Which actions are possible or stimuli received depends also upon the environment and so the agent is situated, interacting with a specific context. Finally, that environment plays an active role in cognition, not just providing stimuli reciprocating action but storing information (through the arrangement of objects, the written word, other agents and so on), and so cognition is said to extend into the environment. 
	
\section{Semiotics and Symbol Emergence}
    \label{semiotics}
	Given the enactive model, what then is the basic unit of cognition? That cognition is the manipulation of abstract symbols is a 20th century notion, prior to which theories were more concerned with sensorimotor stimuli. In order to ground the abstracted symbolism of the 20th century in perception, Barsalou proposed the perceptual symbol system \autocite{barsalou_1999} in which symbols represent states of sensors and actuators. In a computer, input received from a sensor would take the form of a bit string, a bit representing the smallest detectable difference between any two sensory states. In a human, without asserting anything about the form perceptual states might take, this is simply the notion that there must exist some neural representation of physical input, called a perceptual symbol. Perceptual symbols that coincide form a perceptual state, some small subsets of which are selectively extracted and stored in memory. Over time many perceptual states may correspond to the same stored memory. 
	
	A perceptual symbol system is well suited to an enactive model of cognition. Embodiment is necessitated by definition, as to record a perceptual state it is necessary to specify the sensorimotor apparatus which produces it. Cognition extends into the environment in that a perceptual state is the direct result of the interaction between the environment and the sensorimotor apparatus, and so to modify the environment is to modify all subsequent perceptual states. Through this interaction cognition is enacted, embedded within a particular part of the environment.
	
	Importantly, perceptual symbols address the symbol grounding problem, in that they are physically, tangibly real. While there might be some uncertainty as to how this manifests in a biological brain, in the context of a computer a symbol is easily understood as an arrangement of transistors that together form a discrete bit, connected to a larger system which is also tangibly real. Embodied, such symbols represent only themselves. For example, if a sensor attached to a computer constantly outputs to a specific address space, then that memory is the sensory stimuli it represents.  
	
	To connect these low level perceptual states to abstract symbol systems, we look to symbol emergence \autocite{SECD}, 
	the study of how such a high level abstract symbol system may emerge. Emergence is the process of complex, often incomprehensible phenomena arising from the interaction of simple low level systems. The arguably best known example of this is cellular automata, in which an arrangement of discrete cells interact over time using simple rules to produce often dazzlingly intricate patterns. A small change in initial conditions may result in drastic changes to the emergent phenomena. In the same manner, symbol emergence seeks to understand how the enaction of cognition may give rise to an abstract symbol system.
	
	First, an abstract symbol (not a perceptual symbol) is defined as a triadic relationship between a sign, an interpretant and an object, drawing upon Peircean semiotics. Naively we might think of symbols as dyadic and static. We must, in order to facilitate communication, particularly in fields like math where anything less than an exact and immutable definition renders a symbol useless. In the dyadic case a symbol is merely a sign and a referent to which that sign refers. The dyadic structure might describe what is, but not how that came to be the case. 
	
	In the triadic structure the sign, for example the spoken word ``horse", is interpreted by an agent who possesses memories regarding the referent, a horse. The interpretant is the effect of the sign upon the agent which, in the context of the perceptual symbol system above, is the subset of perceptual symbols selectively extracted from various perceptual states. The sign ``horse" is just more sensorimotor information, in this case just a short burst of sound, but through association in memory it induces the sight, smell, emotion and so on associated with experience of the referent (the horse). If an agent observes a horse, the interpretant evokes the sign ``horse", just as the sign evokes the referent. In other words, the interpretant links the sign and referent.
	
	To learn natural language, without knowing what a sign is, an agent will associate the signs employed by other agents in an experience, with that experience. If this happens consistently, the agent can begin to mimic that sign to evoke memories of perceptual states in other agents and so communicate with them. Anecdotally, this might go some way to explaining why it is common for people to attempt to communicate things for which they lack a sign through mimicry of the referent itself (for example, communicating a song for which the name is not known by humming its tune). Even without mimicry, it seems a reasonable conjecture that even an agent raised in isolation might develop its own symbol system, scrawling markings on the environment as reminders to communicate with its future self. As humans we often use language to facilitate thought. While it is possible that an agent might think in perceptual symbols, for example a musician composing music by simulating its performance within their mind, even an agent raised in isolation might ``think" in shorthand, defining its own arbitrary symbol system in order to simplify planning. Communication using symbols requires a community of agents, at least roughly, to share the same signs, referents and interpretants. It is through the interaction of an embodied agent with its environment, including but not limited to other agents with which a shared symbol system or language facilitating communication is desired, that a symbol system might emerge. 
	
	Symbol emergence in robotics seeks to emulate this process so that a robotic agent might not only arrange words to mimic human speech, as is the case with natural language models like GPT-3, but know what those words mean. While the technical specifics are beyond the scope of this paper, for the skeptical reader it is worth noting that this line of inquiry is rapidly bearing fruit \autocite{SER}.
	
	If an agent can be constructed that not only mimics but understands natural language, an agent for which words evoke memories of perceptual states to give them meaning, then that agent can through both experience and testimony be taught complex concepts, such as what it is to be ethical.
	
	\section{Learning Meaning to Understand Ethics}
	\label{ethical_goal}
	Section \ref{intent} described how an ethical framework can be constructed as a concept and a category, defining a concept as an intensional definition (a sentence), and a category as an extensional definition (a set). From the intensional concept the extensional category stems. If the agent learns the concept and then treats that concept as a subgoal of every goal, then that agent has ethical intent. The agent may obtain the concept by finding the \textit{weakest}, most general sentence necessary and sufficient to imply the ostensive definition given the narrow subset of situations to which that ostensive definition pertains. If it is also the case that the ostensive definition is sufficiently representative of the extensional definition (a complex question beyond the scope of this paper), then that sentence will also be necessary and sufficient to imply the extensional definition and subsequently qualify as the intensional definition of the entire category. 
	
	This process is not limited to ethics, however. If concept and category are defined as a sentence and a set as described in section \ref{intent}, as intensional and extensional definitions, then every intensional definition may be obtained from a sufficient ostensive definition, and every extensional definition from an intensional definition. Intuitively, sufficient means any variety of examples from which it is at least possible to learn a specific intensional definition, if not obvious. To say an ostensive definition is not sufficient means it does not imply the specific intensional definition we want, though it may imply a similar one (which would in turn facilitate construction of a slightly different extensional definition containing the ostensive one). Of course, depending on what is being defined, the number and variety of examples required to form an ostensive definition capable of implying such a concept might vary wildly. For example, in the case of random noise there is no meaningful difference between the concept and category because there does not exist any underlying cause and effect relationship which might be exploited, in which case any ostensive definition sufficient to imply the concept \textit{is} the extensional definition, and the concept would simply be the disjunction of all examples in the category. Fortunately, as we're concerned primarily with human ethics and symbol systems, which are far from random noise, this is not an issue.
	
	Natural language tends to contain a lot of ambiguity, with words having more than one meaning depending upon context. Moreover, as frequently pointed out in Judea Pearl's research on causality, for an agent to posses genuine understanding of any topic, it must be able to entertain counterfactuals and ask ``what if $x$ had happened instead of $y$" to distinguish the cause of an effect from correlation \autocite{bookofwhy}. Conventional machine learning methods, as exemplified by GPT-3, make little or no attempt to do this \autocite{gpt}. Symbol emergence in robotics research has primarily focused on models such as latent dirichlet allocation (LDA) in which symbols are latent variables, the meaning of which is learned through unsupervised clustering of multimodal sensorimotor data \autocite{SER}. However, while this is demonstrably effective in associating multimodal sensorimotor stimuli to form symbols, it doesn't really explain what's happening beyond parameter fitting and clustering. Rather than leaping head first into experimenting with possible means to an end, defining an algorithm that learns, it might be better to first examine the desired end, to better define what would constitute success in this entire endeavour.
	
	In a conventional computer a perceptual state is the state of its component parts, the binary values stored at various addresses. An instruction set, written in the physical arrangement of logic gates, interprets the binary code stored at such an address, resulting in a change in perceptual state. 
	
	A perceptual symbol system $\mathcal{PSS}$ must be capable of expressing every aspect of this process, the addresses and values of bits as well as the logic that underpins any computation. The implementation of an instruction set architecture specifying the meaning of machine code might be viewed as a sentence written in a $\mathcal{PSS}$ (machine code language then being a level above a $\mathcal{PSS}$). Just as we described the category of ethical behavior in section \ref{intent}, machine code could then be thought of as a set of situations $\mathcal{S}$ and a set or responses $\mathcal{R}$. A situation is a sentence written in the $\mathcal{PSS}$, one for every variation of every instruction (for example ``jump to address \#" where ``\#" is an integer), likewise for a response which describes the effect of having executed an instruction. For a specific instruction set architecture, there exists a set $\mathcal{D} = \{(s, r) \in \mathcal{S} \times \mathcal{R} : (s, r) \ is \ a \ part \ of \ the \ instruction \ set\}$. There exists a concept $c$, a sentence in $\mathcal{PSS}$ such that $c$ is true if and only if $(s, r)$ is in $\mathcal{D}$. If $c$ were then used as a goal, a rational agent would act according to the instruction set. $c$ describes the instruction set.
	
	Any higher level language $\mathcal{L}$ could be specified in this manner using the lower level $\mathcal{PSS}$. This is analogous to the way logic is physically encoded through the arrangement of gates to specify machine code, which is used to specify assembly, then compiled languages and finally high level interpreted languages such as Python. 
	
	Assume $\mathcal{L}$ is a symbol system encapsulating all natural language we wish an agent to learn, in order to communicate with humans and construct an ethical framework. $\mathcal{L}$ includes not just written and spoken words but facial expressions, cultural cues and so on. The purpose of $\mathcal{L}$ is to convey meaning, not just mimic utterances. It is, in effect, a means of encoding and decoding frequently occurring subsets of perceptual states for efficient transmission between agents using every available mode of interaction (because it is grounded in the multimodal $\mathcal{PSS}$). Signs and referents are deeply integrated with context as a result.
	Because $\mathcal{L}$ is to be specified in a $\mathcal{PSS}$, a symbol in $\mathcal{L}$ is defined a little differently from the aforementioned triadic Piercean semiosis. The signs, interpretants and referents of a symbol in $\mathcal{L}$ are each described in the $\mathcal{PSS}$. The agent may recall any stimuli associated with a symbol and then transmit a sign of that symbol, or observe either a sign or referent of that symbol and recall past experiences associated with that symbol. The relevant distinction is between a situation and a response. A sign is just a referent that the agent is capable of transmitting in some manner, and  ideally should be sufficient to unambiguously identify a specific symbol. 
	
	A concept may be arbitrarily narrow or broad in its scope, so just as an entire language like machine code could be described in $\mathcal{PSS}$ as a concept $c$ and category $\mathcal{D}$, so may a solitary symbol in $\mathcal{L}$. For a solitary symbol, there exists a set of situations $\mathcal{S}$, each situation made up of observations signs and referents in context. There exists a set of all possible responses $\mathcal{R}$ which is all possible means of multimodal signalling and of recalling memories.
	
	As words in natural language tend to be ambiguous, not all words associated with a symbol are appropriate for all situations (for example the French and Japanese languages may share symbols, but communicate them using different signs), and words may be parts of signs used to refer to different symbols in different contexts. A situation must include who is speaking or being spoken too, as well as placement in a sentence and so on. To understand what someone means by a sentence, the task is to infer from a situation which symbol is intended, to respond by accessing memories associated with that symbol. Conversely to communicate a message in an ambiguous, contextually dependent language the task is to convey the desired symbol using a response. To respond with a particular sign is appropriate if it is likely to convey the intended meaning (symbol), meaning that the speaker is making an educated guess as to what the listener's interpretant is. The members of a given population will share interpretants, signs and referents to some extent, otherwise communication would be impossible \autocite{SECD}. For every symbol there exists a set $\mathcal{D} = \{(s, r) \in \mathcal{S} \times \mathcal{R} : (s, r) \ is \ appropriate\}$. This is the category of all situation-response pairs which succeed in conveying or interpreting a specific symbol, taking into account cultural context, who is speaking or being spoken too and so on. Just as with the concept of ethics described in section \ref{intent}, for every symbol there exists a concept $c$ written in $\mathcal{PSS}$ such that given any pair $(s, r)$, $c$ is true if and only if $(s, r) \in \mathcal{D}$. Given a situation $s$, $c$ may be used to abduct $(s, r)$ appropriate to convey the intended symbol. Alternatively, if the agent observes stimuli associated with a symbol, then it responds to that situation by accessing memories associated with that symbol. 
	
	Constructing a symbol in this manner also means signs may be associated with other signs, and referents with referents. Symbols may therefore vary in scope of meaning, encapsulating either many or few situation and response pairs, either many different referents and signs or as few as one referent with one sign as with a specific instruction in machine code. Unlike machine code, human communication tends to be inexact. The concept of a symbol in human language is weaker, less specific. 
	When observing another individual speak, the agent may abduct which symbols may be intended by the speaker by finding all the symbols in $\mathcal{L}$ for which the observed $(s, r)$ was appropriate. In this approximate manner an agent may interpret the signs transmitted by another. 
	
	To explain how this compares with triadic Piercean semiosis (introduced in section \ref{semiotics}), the concept $c$ of a symbol is performing the role of the interpretant, and the category $\mathcal{D}$ is the set of all appropriate pairings associated with the symbol. It is important to note that $c$ is the same concept whether the task is to interpret or communicate a symbol. 
	
	As before, through observing and interacting with humans, the agent may learn a symbol from an ostensive definition $\mathcal{O}_\mathcal{D} \in \mathcal{D}$ by finding the weakest, most general sentence $c$ in $\mathcal{PSS}$ which is both necessary and sufficient to imply $\mathcal{O}_\mathcal{D}$ in a subset of possible situations $\mathcal{O}_\mathcal{S}$ to which the ostensive definition pertains. Remember, it is important that it is the weakest necessary and sufficient sentence as this is the most likely to also be necessary and sufficient to imply $\mathcal{D}$ given $\mathcal{S}$. As with the concept of ethics, the interpretant $c$ of a symbol is most likely to be shared by other agents if it is the weakest, most general $c$ necessary and sufficient to imply $\mathcal{O}_\mathcal{D}$, because it will encapsulate the greatest breadth of experience. By learning several symbols in this manner the agent can construct a rudimentary vocabulary within $\mathcal{L}$, at which point it can learn new symbols by observing communication between humans in $\mathcal{L}$ for which it understands some of the signs (meaning it correctly abducts the appropriate symbols from each pair $(s, r)$ observed). Through extensive observation and interaction, the agent may eventually develop a rich vocabulary capable of understanding what is meant by a human.
	
	Importantly, such an agent will be capable of using counterfactuals, of knowing what responses would have also been appropriate given preceding communication, and what other messages might have been conveyed. It does not learn a function, but instead the spoken and unspoken rules of human communication within populations. These symbols are learned in context, they have meaning in terms of multimodal sensorimotor experience of specific people and places. In doing so it can learn what it is to be ethical, not just in terms of testimony from humans but observation of their actions. The responses or descriptions of responses undertaken by others share abstract symbols with the actions of the agent itself, meaning it can understand that if it was unethical for a person in a particular situation to choose a particular response, then it would be unethical to choose that pair $(s, r)$ itself.
	
	\section{The Mirror Symbol Hypothesis}
	\label{mirrorsymbol theory}
	Mirror neurons in the human brain discharge both when executing a specific motor act (such as grasping), and when observing another individual performing that same act \autocite{RIZZOLATTI2015677}. This occurs in a multi-modal sense, for example where mirror neurons activate both when hearing a phoneme, and when using muscles to reproduce that phoneme. Such neurons are suspected to play a role in the inference of intent, because the activation of mirror neurons reflects not just the actions of those we're observing, but the context and anticipated purpose for which actions are taken. Most importantly, mirror neurons also reflect the emotions of those we observe; the observation of laughter, disgust and so forth triggers the same in ourselves. In short, mirror neurons may facilitate empathy \autocite{RIZZOLATTI2019}.
	
	A symbol of $\mathcal{L}$ which represents some characteristic of a pair $(s, r)$ is present both when $(s, r)$ is enacted by the agent, and also when observing another individual respond to a situation in the manner of $(s, r)$. The concept $c$ is the same for both receiving and transmitting a symbol. $c$ is true if and only if the agent observes a sign \textit{or} referent associated with the symbol in question, or when the agent transmits a sign to communicate the symbol. This shared purpose of interpretants may allow the agent to empathise, because an experience of its own is associated with a specific symbol, which is also associated with the observation of another agent experiencing something that symbol represents. In both of these cases $c$ is true, linking the present to past experience.
	
	What we have described up until now is an agent which constructs an intensional definition from an ostensive definition. From that intensional definition an extensional definition could be constructed, and it follows that an objective function exists which is maximised by choosing only those pairs $(s, r) \in \mathcal{D}$. Conversely, an agent may be given only an objective function which scores $(s, r)$ pairs. That agent initially has no knowledge of any ostensive definition, but there exists some extensional definition which maximises that objective function. By interacting with the world and attempting to maximise this objective function, that agent could construct an ostensive definition, followed by an intensional definition, followed by the extensional definition for which that function is maximised. Thus, to learn a concept an agent may be provided either an ostensive definition, or an objective function. What follows assumes an agent can be given objective functions closely approximating those that seem to be built into human sensorimotor systems; pleasure, pain, hunger and so forth. How such objective functions come to exist is beyond the scope of this paper, but is discussed in section \ref{conclusion} in relation to the emergence of normativity.
	
	Empathy is a complex phenomenon and the definition adopted here is not intended to encompass all that might be associated with the term. There are many competing accounts of what empathy entails \autocite{sep-empathy, zahavi_2014}. We have built upon Hoffman’s view \autocite{hoffman_2000} of empathy which includes both shallow and higher-order cognitive processing. In the context of this paper to empathise is to infer what another individual is experiencing, what that individual is likely to want (is compelled to seek by its goals or objective functions)\footnote{Using the term ``want" in a very restrictive sense.} given that experience, relating that to one's own similar past experiences to feel something of that want. As we have defined intent, an agent can come to possess intent by constructing a concept from an ostensive definition (inferring purpose from observation), which can in turn be created using objective functions. Intent refers to either $c$ or some aspect of it. To clarify the latter; the scope of a concept being arbitrary, intent in the context of one specific pair $(s, r)$ might refer to those parts of $c$ which motivated that choice (a similar notion to specific neurons activating in response to specific stimuli). If $c$ serves as a goal constraint, an agent possessed of overall intent $c$ can explain another individual's observed behaviour by representing it internally as a pair $(s, r)$ and assuming that other individual pursues similar overall goals. Then, to say the agent infers the intent of another individual is to say it imbues a specific instance of their behaviour with presumed intent in terms of its own goals (and any objective functions used to construct that goal). It knows why it would choose a response, assumes the other agent is the same, and by representing the other agent's behaviour internally as $(s, r)$ may in some way feel compelled by its own objective functions. Such a definition requires agents labour under similar compulsions, and have comparable past experiences, for empathy or the correct inference of intent to be at all possible (otherwise they would each construct very different intents $c$). Assuming that is the case, whether this results in empathy or a form of emotional contagion depends upon the ability of the agent to distinguish between itself and the other agent; other-directed intentionality \autocite{zahavi_2014}.
	
	To further illustrate the idea, when the agent observes another individual experiencing an event, it also recalls its own similar experiences. Because the concept $c$ is pursued as a goal, it imbues stimuli with significance in terms of that goal. It assigns intent to observed behaviour where intent is a goal as described in section \ref{intent}, or some aspect of it (the scope of a concept being arbitrary). If the intent of another agent being observed is similar to $c$, then the significance with which $c$ imbues that other agent's behaviour may resemble the intent of that other agent. For example, upon witnessing an individual suffering the agent will recall experiences of suffering, and infer that the suffering individual wishes the suffering to cease (if that is what the compulsions informing the construction of $c$ would dictate). This is because referents are associated with referents as well as signs by the interpretant, and those referents indicate something of significance to a goal. 
	
	In any case, none of this can be achieved without the ability to give an agent objective functions (and subsequently experiences) similar to those humans labour under. However, even if an agent were not given objective functions analogous to human feelings, that agent might still observe the reactions of humans, the observable extent of their pain or pleasure, and associate those with symbols, so long as those phenomena are of at least some relevance to the agent's objective functions. Though we are not making any claims about consciousness, this would at least allow the agent to infer something of what a human it observes is experiencing, which could then be used to iteratively improve upon the objective functions compelling the agent so that they more closely reflect the human condition. Compelled by such objective functions, it may recall experiencing a feeling when it observes a human experiencing that feeling. Of course, constructing such objective functions seems a very difficult proposition.
	
	\section{Concluding Remarks and Comparison to Related Work}
	\label{conclusion} 
	The preceding specified an agent that can understand rather than merely mimic natural language, infer and possess intent, cope with ambiguity, learn social and moral norms and empathise with other agents in terms of its own past experience.
	
	The remainder of this paper provides context and comparison with existing research. 
	
	The emergence of normativity is not resolved here, and it is debatable whether sensorimotor and semiotic activity is sufficient for such emergence. Arguably, normativity is given top-down rather than bottom-up. To address this, future research might explore how normativity may come about as a result of adaptive evolution \autocite{sep-morality-biology}, through a long empirical process \autocite{bennett_2021_c} emerging naturally rather than being imposed \autocite{bennett_2021}. Such an account of ethics must address classic issues such as Hume's is-ought problem and Moore's naturalistic fallacy \autocite{sep-moral-non-naturalism}, and may benefit from the integration of counterfactual language \autocite{lewis_1963, bookofwhy}.
	
	The unintended consequences of simplistic or prescribed moral frameworks for artificial intelligence are a popular topic \autocite{asimov} in speculative fiction in some part because the unwritten rules of human interaction are based upon ever changing social and moral norms. If an artificial intelligence act in accordance with those norms it must learn, represent and reason with those norms. It must be explicitly ethical, able to navigate ambiguity and ethical dilemmas unanticipated by its designers \autocite{yu2018building, Scheutz_2017}. 
	An agent meeting the above specification will learn a concept of ethics which will then constrain its behaviour if included in its goals. This concept will be learned from the humans with which the agent interacts, and coincide the ethical beliefs of those humans. By identifying the weakest constraints $c$ as described in section \ref{intent}, it will adopt the most commonly held ethical beliefs where they exist within any particular subgroup, yet retain flexibility to learn the beliefs of a single individual where that individual deviates from the norm, or when a situation presents a moral dilemma. It may learn exceptions to any rule and account for context. It will avoid the pitfalls of simplistic prescribed ethical frameworks popular in speculative fiction, such as the work of Isaac Asimov concerning his Three Laws of Robotics (bearing in mind the aforementioned risk of learning undesirable ethical norms through exposure to bad actors early in its existence).

	Cause and intent, both of which are central to human legal frameworks, have been raised as issues that must be solved if an artificial intelligence is to be considered ethical \autocite{Bathaee2018TheAI}.
	Unlike existing deep learning based natural language models such as GPT-3 \autocite{gpt}, the signs communicated by this agent have meaning grounded in multimodal sensorimotor stimuli, and it possesses intent in the form of an explicitly ethical goal. An ethical agent must also be able to explain and justify its decisions \autocite{Scheutz_2017}. An agent meeting the above specification can express in a comprehensible manner the explicit criteria upon which it judges a response ethical or not. Even if it is too complex to be easily interpretable, and though it may exhibit somewhat inconsistent behaviour because it is a learning agent, it is trustworthy because it possesses and can communicate its intent. Actions can be explained in the context of this intent.
	Because it can explain itself, and because its decisions are based upon identifiable constraints, it addresses the issues of transparency and explainability \autocite{BARREDOARRIETA202082}, which are often cited as necessary conditions for ethical AI. Though easily interpretable ML methods such as graphical models already exist, a human expert is still required to analyse and explain behaviour. In contrast, the agent specified here would be able to actively explain itself, respond to questions in a manner comprehensible to a layperson.
	
	The truth value of $c$ defined here, the weakest necessary and sufficient sentence which implies an ostensive definition, will ignore merely correlated information and account only for what is absolutely necessary to determine if a pair $(s, r)$ is a member of a category. As such it addresses the problem of causality \autocite{bookofwhy, BARREDOARRIETA202082, Bathaee2018TheAI} in that an agent that learns $c$ will not make judgements based on correlated information if causal relationships are observable. It does not learn a function, but the written and unwritten rules by which the humans with which it interacts judge membership of a category. Algorithmic bias resulting from correlations would be eliminated. Because it can entertain counterfactuals, it can adjust its decisions given additional information. For example, if it were handling administrative procedures, it would be able to holistically consider extenuating circumstances and appeals.
	
    With mirror symbols, it may empathise in the sense defined in section \ref{mirrorsymbol theory}.  While empathy is not explicitly listed as a necessary condition for ethical artificial intelligence in the aforementioned literature, it is arguably necessary if an agent is to comprehend and comply with social and moral norms. Humans make decisions not with logic alone, but with sentiment. To comprehend sentiment it must infer what is intended by a human, rather than what is explicitly stated. A concept $c$ that resembles human purposes may facilitate this, imbuing signs with human-like intent, what a human means rather than says inasmuch as such a thing exists in the agent's own experience. 

    Finally, the prospect of Artificial General Intelligence (AGI) becoming an existential threat to humanity is often cited as motive for the study of ethical AI \autocite{yu2018building}, but is less often considered as a requirement for ethical AI. Linguistic fluency, intent, empathy and an understanding of context are certainly not found in the narrow AI commonly employed today. Subsequently, narrow AI is arguably incapable of being ethical \autocite{hew2014}. Linguistic fluency and intent may be qualities of the villainous AGI of speculative fiction, but empathy and the flexibility necessary to deal with exceptions to rules are usually not. Yet its arguable that all of these qualities are necessary for both ethical AI and AGI \autocite{bennett_2021}. As this specification may be applied to any concept, not just ethics, it may be considered a blueprint for the development of an Ethical AGI.

    \section*{Acknowledgements}
    We would like to acknowledge and thank the reviewers for their thoughtful and detailed feedback, which improved this paper substantially.









\ifCLASSOPTIONcaptionsoff
  \newpage
\fi


\printbibliography

@article{gpt,
	journal = {Minds and Machines},
	title = {GPT-3: Its Nature, Scope, Limits, and Consequences},
	year = {2020},
	author = {Luciano Floridi and Massimo Chiriatti},
	pages = {1-14}
}

@Inbook{Stilgoe2020,
    author="Stilgoe, Jack",
    title="Who Killed Elaine Herzberg?",
    bookTitle="Who's Driving Innovation? New Technologies and the Collaborative State",
    year="2020",
    publisher="Springer International Publishing",
    address="Cham",
    pages="1-6"
}

@electronic{hp2019,
	Address = {New York},
	Author = {Visser, Nick},
	Db = {ProQuest One Academic},
	Journal = {The Huffington Post},
	La = {English},
	Publisher = {AOL Inc.},
	Title = {Uber Not Criminally Liable After Self-Driving Car Killed Woman: Local Prosecutor: The death of 49-year-old Elaine Herzberg was the first known fatality linked to an autonomous vehicle in the United States.},
	Year = {2019}
}

@article{BARREDOARRIETA202082,
    title = "Explainable Artificial Intelligence (XAI): Concepts, taxonomies, opportunities and challenges toward responsible AI",
    journal = "Information Fusion",
    volume = "58",
    pages = "82-115",
    year = "2020",
    author = "{Alejandro et al.} {Barredo Arrieta}"
}

@incollection{RIZZOLATTI2019,
    title = "The Mirror Neuron Mechanism",
    booktitle = "Reference Module in Neuroscience and Biobehavioral Psychology",
    publisher = "Elsevier",
    year = "2019",
    author = "Giacomo Rizzolatti and Maddalena Fabbri-Destro and Marzio Gerbella"
}

@incollection{RIZZOLATTI2015677,
    title = "Action Understanding",
    editor = "Arthur W. Toga",
    booktitle = "Brain Mapping",
    publisher = "Academic Press",
    address = "Waltham",
    pages = "677-682",
    year = "2015",
    author = "Giacomo Rizzolatti"
}

@book{Hutter2005UniversalAI,
	author        = "Hutter, Marcus",
	title         = "{Universal artificial intelligence: Sequential decisions based on algorithmic probability}",
	publisher     = "Springer",
	year          = "2005"
}

@article{Scheutz_2017, 
	title={The Case for Explicit Ethical Agents},
	volume={38}, 
	journal={AI Magazine},
	author={Scheutz, Matthias}, year={2017}, 
	pages={57-64} 
}

@article{Bathaee2018TheAI,
	title={The Artificial Intelligence Black Box and the Failure of Intent and Causation},
	author={Yavar Bathaee},
	journal={Harvard Journal of Law and Technology},
	year={2018},
	volume={31},
	pages={889}
}

@article{yu2018building,
	title={Building ethics into artificial intelligence},
	author={Yu, Han and Shen, Zhiqi and Miao, Chunyan and Leung, Cyril and Lesser, Victor R and Yang, Qiang},
	journal={arXiv preprint arXiv:1812.02953},
	year={2018}
}

@incollection{CARPENTER202075,
	title = "Chapter 4 - Kill switch: The evolution of road rage in an increasingly AI car culture",
	editor = "Richard Pak and Ewart J. {de Visser} and Ericka Rovira",
	booktitle = "Living with Robots",
	publisher = "Academic Press",
	pages = "75-90",
	year = "2020",
	author = "Julie Carpenter",
}

@book{bookofwhy,
	author = {Pearl, Judea and Mackenzie, Dana},
	title = {The Book of Why: The New Science of Cause and Effect},
	year = {2018},
	publisher = {Basic Books, Inc.},
	edition = {1st}
}

@article{barsalou_1999, 
	title={Perceptual symbol systems}, 
	volume={22}, 
	journal={Behavioral and Brain Sciences}, 
	publisher={Cambridge University Press},
	author={Barsalou, Lawrence W.}, 
	year={1999}, 
	pages={577-660}
}

@InCollection{sep-intention,
	author       =	{Setiya, Kieran},
	title        =	{{Intention}},
	booktitle    =	{The {Stanford} Encyclopedia of Philosophy},
	editor       =	{Edward N. Zalta},
	year         =	{2018},
	edition      =	{Fall 2018},
	publisher    =	{Metaphysics Research Lab, Stanford University}
}

@InCollection{sep-definitions,
	author       =	{Gupta, Anil},
	title        =	{{Definitions}},
	booktitle    =	{The {Stanford} Encyclopedia of Philosophy},
	editor       =	{Edward N. Zalta},
	year         =	{2019},
	edition      =	{Winter 2019},
	publisher    =	{Metaphysics Research Lab, Stanford University}
}

@article{SER,
	author = {Tadahiro Taniguchi and Takayuki Nagai and Tomoaki Nakamura and Naoto Iwahashi and Tetsuya Ogata and Hideki Asoh},
	title = {Symbol emergence in robotics: a survey},
	journal = {Advanced Robotics},
	volume = {30},
	number = {11-12},
	pages = {706-728},
	year  = {2016},
	publisher = {Taylor & Francis}
}

@MISC{Harnad90thesymbol,
	author = {Stevan Harnad},
	title = {The Symbol Grounding Problem},
	year = {1990}
}

@MISC{asimov,
	author = {Isaac Asimov},
	title = {Runaround},
	year = {1942}
}

@ARTICLE{SECD,
	author={Tadahiro {Taniguchi} and Emre {Ugur} and Matej {Hoffmann} and Lorenzo {Jamone} and Takayuki {Nagai} and Benjamin {Rosman} and Toshihiko {Matsuka} and Naoto {Iwahashi} and Erhan {Oztop} and Justus {Piater} and Florentin {Wörgötter}},
	journal={IEEE Transactions on Cognitive and Developmental Systems}, 
	title={Symbol Emergence in Cognitive Developmental Systems: A Survey}, 
	year={2019},
	volume={11},
	number={4},
	pages={494-516}
}

@book{mindinlife,
	author = {Thompson, Evan},
	year = {2007},
	title = {Mind in Life. Biology, Phenomenology and the Sciences of Mind},
	volume = {18},
	journal = {Journal of Consciousness Studies}
}

@ARTICLE{Newell_physicalsymbol,
	author = {Allen Newell},
	title = {Physical symbol systems},
	journal = {Cog. Sci},
	year = {1980},
	pages = {135-183}
}

@article{hew2014,
	Author = {Hew, Patrick Chisan},
	Journal = {Ethics and Information Technology},
	Number = {3},
	Pages = {197-206},
	Title = {Artificial moral agents are infeasible with foreseeable technologies},
	Volume = {16},
	Year = {2014},
}

@article{lewis_1963,
	Author = {Lewis, David},
	Journal = {Journal of Philosophy},
	Title = {Causation},
	Year = {1973}
}

@InCollection{sep-empathy,
	author       =	{Stueber, Karsten},
	title        =	{{Empathy}},
	booktitle    =	{The {Stanford} Encyclopedia of Philosophy},
	editor       =	{Edward N. Zalta},
	year         =	{2019},
	edition      =	{{F}all 2019},
	publisher    =	{Metaphysics Research Lab, Stanford University}
}

@book{hoffman_2000, place={Cambridge}, title={Empathy and Moral Development: Implications for Caring and Justice}, publisher={Cambridge University Press}, author={Hoffman, Martin L.}, year={2000}}

@misc{bennett_2021_b,
  title={The Solutions to Any Task},
  author={Michael Timothy Bennett},
  howpublished = {PhD Thesis Manuscript},
  year={2021}
}

@misc{bennett_2021,
  title={Symbol Emergence and The Solutions to Any Task},
  author={Michael Timothy Bennett},
  howpublished = {Under consideration},
  year={2021}
}

@misc{bennett_2021_c,
  title={The Artificial Scientist: Logicist, Emergentist, and Universalist Approaches to Artificial General Intelligence},
  author={Michael Timothy Bennett and Yoshihiro Maruyama},
  howpublished = {Under consideration},
  year={2021}
}

@article{zahavi_2014,
    author = {Zahavi, Dan},
    year = {2014},
    title = {Empathy and Other-Directed Intentionality},
    volume = {33},
    journal = {Topoi}
}

@InCollection{sep-morality-biology,
	author       =	{FitzPatrick, William},
	title        =	{{Morality and Evolutionary Biology}},
	booktitle    =	{The {Stanford} Encyclopedia of Philosophy},
	editor       =	{Edward N. Zalta},
	howpublished =	{\url{https://plato.stanford.edu/archives/spr2021/entries/morality-biology/}},
	year         =	{2021},
	edition      =	{{S}pring 2021},
	publisher    =	{Metaphysics Research Lab, Stanford University}
}

@InCollection{sep-moral-non-naturalism,
	author       =	{Ridge, Michael},
	title        =	{{Moral Non-Naturalism}},
	booktitle    =	{The {Stanford} Encyclopedia of Philosophy},
	editor       =	{Edward N. Zalta},
	howpublished =	{\url{https://plato.stanford.edu/archives/fall2019/entries/moral-non-naturalism/}},
	year         =	{2019},
	edition      =	{{F}all 2019},
	publisher    =	{Metaphysics Research Lab, Stanford University}
}

%




\end{document}